\documentclass[runningheads,a4paper]{llncs}

\usepackage{amssymb}
\usepackage{graphics}
\usepackage{graphicx}
\usepackage{prettyref}
\usepackage{bm}
\usepackage{amsmath}
\usepackage{xspace}
\usepackage{multirow}
\usepackage{float}

\usepackage{lipsum}  % fill text
\usepackage{siunitx} % units
\usepackage[numbers,sort&compress,sectionbib]{natbib} % sorted cites
\usepackage[hidelinks]{hyperref} % hyperlink references
\usepackage[capitalize]{cleveref} % clever references
\usepackage[tableposition=top,font={small}]{caption}
\usepackage{subcaption}
\usepackage{tabularx,booktabs}
\usepackage{wrapfig} % wrapping figures
\usepackage{tikz}
\usepackage[bottom]{footmisc} % footnotes at bottom
\def\check{\pmb{\tikz\fill[scale=0.3](0,.35) -- (.25,0) -- (1,.7) -- (.25,.15) -- cycle;}}
\usepackage{forest}

\usepackage{url}

\let\oldmarginpar\marginpar
\renewcommand\marginpar[1]{\oldmarginpar[\raggedleft\footnotesize #1]%
{\raggedright\footnotesize #1}}

\newcommand{\tsmcomment}[2]{}
\newcommand{\vmcomment}[2]{}
\newcommand{\skcomment}[2]{}
\newcommand{\aacomment}[2]{}
\newcommand{\wgcomment}[2]{}

\newcommand{\ms}[1]{\SI{#1}{\metre\per\second}}
\newcommand{\Hz}[1]{\SI{#1}{\hertz}}

\newcommand{\hr}[1]{\SI{#1}{\hour}}
\newcommand{\Kg}[1]{\SI{#1}{\kilogram}}
\newcommand{\km}[1]{\SI{#1}{\kilo\metre}}
\newcommand{\met}[1]{\SI{#1}{\metre}}
\newcommand{\inertiam}[1]{\SI{#1}{\kilogram\per\square\metre}}

\newcommand{\numtotalflights}{168}

\newcommand{\maxspeed}{\ms{7.0}}
\newcommand{\numtrajectories}{17}
\newcommand{\numenvironments}{5}

\usepackage{filecontents}

\begin{document}
\mainmatter  % start of an individual contribution

\title{The Blackbird Dataset:\\A large-scale dataset for UAV perception in aggressive flight}
\titlerunning{A large scale visual inertial-dynamical dataset for agile UAV perception}

\author{Amado Antonini\inst{1} \and Winter Guerra\inst{1} \and Varun Murali\inst{1} \and Thomas Sayre-McCord\inst{1} \and Sertac Karaman\inst{1}}%
\authorrunning{Antonini et al.}

\institute{
Laboratory for Information and Decision Systems,\\
Massachusetts Institute of Technology
\thanks{Emails: \email{\{amadoa, winterg, mvarun, rtsm, sertac\}@mit.edu}}
% \and
% Army Research Lab
}

\newcommand{\myparagraph}[1]{\textbf{#1.}\xspace}

\toctitle{Dataset for Agile UAV SLAM}
\tocauthor{Antonini et al.}
\maketitle

\begin{abstract}
%   In the past few years, autonomous unmanned aerial vehicle (UAV) technology has received a significant amount of attention.
%   At the same time, visual inertial navigation has shown promising results as a way of achieving this autonomy due to the portability of the sensors it uses, intertial measurement units (IMU's) and cameras.
%   However, significant development is required for the state-of-the-art visual inertial systems to achieve the performance and robustness needed in many applications.
%   This work presents a large-scale dataset that facilitates the development of more robust autonomous UAV navigation systems by providing a great variety of visual, inertial, and dynamical sensor data from an agile quadrotor platform during over \hr{10} of flight.
% The size and variety of the dataset make it a valuable tool for evaluating and testing visual-intertial estimation algorithms.

The Blackbird unmanned aerial vehicle (UAV) dataset is a large-scale, aggressive indoor flight dataset collected using a custom-built quadrotor platform for use in evaluation of agile perception.
Inspired by the potential of future high-speed fully-autonomous drone racing,
the Blackbird dataset contains over 10 hours of flight data from \numtotalflights{} flights over \numtrajectories{} flight trajectories and \numenvironments{} environments at velocities up to \maxspeed{}. 
Each flight includes sensor data from \Hz{120} stereo and downward-facing photorealistic virtual cameras, \Hz{100} IMU, $\sim$\Hz{190} motor speed sensors, and \Hz{360} millimeter-accurate motion capture ground truth.
Camera images for each flight were photorealistically rendered using FlightGoggles \cite{mccord2018ICRA} across a variety of environments to facilitate easy experimentation of high performance perception algorithms.
The dataset is available for download at \url{http://blackbird-dataset.mit.edu/}.

% By providing the same trajectories at multiple speeds, we make it easy to isolate the performance of perception algorithms against flight agility and flight trajectory, and rapidly identify bottlenecks in sensors and algorithms.

% The visual inertial dataset presented here comprises of high rate sensor data--including 3 cameras, an IMU, and rotary encoders--and high accuracy ground truth pose for a  set of 17 indoor flight trajectories, each of which is repeated with increasing agility. 
% By providing the same trajectories at multiple speeds, we make it easy to isolate the performance of perception algorithms against flight agility and flight trajectory, and rapidly identify bottlenecks in sensors and algorithms.

\end{abstract}

% Motivation, Problem Statement, Related Work (one page)
\section{Introduction}
\label{sec:intro} 

Aggressive Unmanned Aerial Vehicle (UAV) flight using visual inertial simultaneous localization and mapping (VI-SLAM) has received increasing attention over recent years \cite{falanga2017aggressive,mccord2018ICRA}.
With the availability of better hardware, aggressive indoor flight maneuvers that were previously only possible using motion capture systems are now becoming achievable using on-board visual inertial state estimation algorithms.
In the near future, it is conceivable that complex high-speed tasks, such as fully-autonomous drone racing, will be possible in realtime.
To aid in the development of these high-performance algorithms, we provide a large scale, high rate, and high accuracy dataset for the improvement and evaluation of VI-SLAM for agile indoor flight.

\begin{table}[tbh!]
    %\begin{table}[b]
\centering
\caption{UAV Visual Inertial Datasets Comparison}
\resizebox{0.9\textwidth}{!}{\begin{minipage}{1.0\textwidth}
{\setlength{\tabcolsep}{1.1em}
  \begin {tabularx} {\textwidth} {ccccc} %{XXXXX}
    \hline
    & EuRoC                    & UPenn fast               & Zurich Urban               &               \tabularnewline
    & MAV\cite{burri2016euroc} & flight \cite{sun2018RAL} & MAV \cite{majdik2017IJRR}  & \textbf{Ours} \tabularnewline
    \hline
  Environments & 2 & 1 & 3 & \textbf{5}\footnote{Additional environments may be rendered using FlightGoggles} 
    \tabularnewline
    Sequences & 11 &  4 & 1 & \textbf{186} 
    \tabularnewline
    Camera &   \Hz{20} &  \Hz{40} &  \Hz{20} &  \textbf{\Hz{120}}
    \tabularnewline
    IMU &  \textbf{\Hz{200}} &  \textbf{\Hz{200}} &  \Hz{10} &  \Hz{100} 
    \tabularnewline
    Motor Encoders & n/a & n/a & n/a &  \textbf{\Hz{\sim 190}} 
    \tabularnewline
    Max Distance & \met{130.9} & \met{700} & \textbf{\km{2}} & \met{860.8} 
    \tabularnewline
    Top Speed & \ms{2.3} & \textbf{\ms{17.5}} & \ms{3.9}\footnote{Instantaneous velocity from GPS} & \maxspeed{} 
    \tabularnewline         
    mm Ground Truth &  (\Hz{100}) \footnote{High accuracy only available for half the sequences} & n/a & n/a &  \textbf{\Hz{360}} 
    \tabularnewline
    \hline		 
  \end{tabularx}}
\label{tab:datasetComparison}
\end{minipage}}
%\end{table}
    \end{table}

\begin{figure}[tb]
    \centering
    \includegraphics[width=0.6\textwidth]{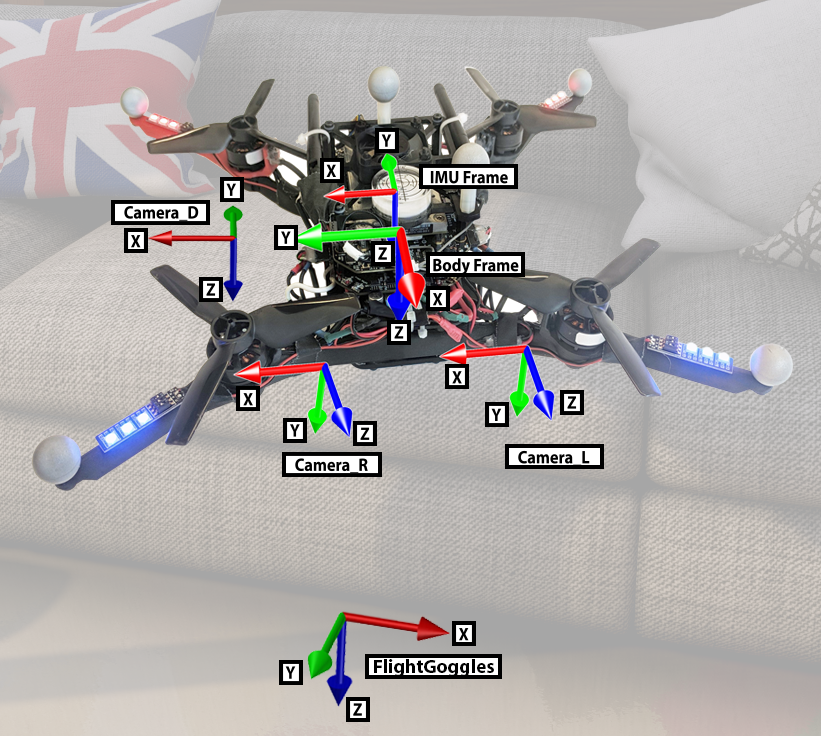}
      \caption[Blackbird]{Coordinate frames in use for this dataset. Note that \texttt{Camera\_D} and \texttt{Body Frame} are coincident, but are translated in the figure for visualization.}
        \label{fig:blackbird}
  \end{figure} 

\myparagraph{Related Work}
As can be seen from \Cref{tab:datasetComparison}, existing UAV datasets focus on either slow speed indoor flight  \cite{burri2016euroc} where high accuracy ground truth is achievable, or outdoor high speed flight  \cite{sun2018RAL,majdik2017IJRR,wang2016} with lower quality ground truth using GPS systems.
Burri et al \cite{burri2016euroc} present the EuRoC MAV datasets, a collection of 11 trajectories with an average speed of ~\ms{1} and highly accurate ground truth for one half of the sequences. 
However, the scenes are not representative of typical indoor environments and are captured using comparatively lower rate cameras.
Sun et al \cite{sun2018RAL} present a fast outdoor flight dataset with the same trajectory at 4 different speeds and GPS ground truth. Although this does allow the evaluation of online algorithms in outdoor settings, it does not provide high quality ground truth or different environments.
The Zurich Urban MAV Dataset is presented by Majdik, Till, and Scaramuzza. It contains \km{2} of visual and inertial data recorded from a tethered UAV flying in an urban setting, but it lacks high-precision ground truth pose.
Wang et al \cite{wang2016} present TorontoCity, a very large UAV dataset with data from multiple perspectives of the city of Toronto captured from different cameras and a LiDAR. TorontoCity focuses on tasks such as segmentation and classification of the environment. 
It, however does not contain inertial information and cannot be used in the context of visual inertial navigation.

% Technical Approach (one page)
\section{Data Collection Setup}
\label{sec:data_collection_setup}

\myparagraph{UAV Platform}
Data was collected using a custom built quadrotor UAV designed for agile autonomous flight, which we call Blackbird (Figure \ref{fig:blackbird}).
The UAV caries an Xsens MTi-3 IMU, custom made optical motor encoders for accurate motor speed measurements, a DJI Snail propulsion system, and a NVIDIA Jetson TX2.
The body of the vehicle is constructed from 3D printed MarkForged Onyx continuous carbon fiber composite. Rubber dampeners are used to mechanically isolate vibrations from the propulsion system from flight sensors.
The physical properties of the quadrotor as well as sensor statistics are shown in \cref{tab:blackbird_phys}.

\begin{table}[tbh!]
	\centering
\caption{Quadrotor characteristics}
\label{tab:blackbird_phys}
{\setlength{\tabcolsep}{1em}
\begin{tabular}{|c|c|l|}
    \hline
    \textbf{Property}         & \textbf{Value}                   & \multicolumn{1}{c|}{\textbf{Description}} \\ 
    \hline
    Mass                      & \Kg{0.915}                       & Mass with battery                         \\
    $I_{xx}$                  & \inertiam{4.9e-2}                & X moment of inertia                      \\
    $I_{yy}$                  & \inertiam{4.9e-2}                & Y moment of inertia                      \\
    $I_{zz}$                  & \inertiam{6.9e-2}                & Z moment of inertia                      \\
    Arm Length                & \SI{0.13}{\metre}                & Center to end of arm                      \\
    $f_x$, $f_y$              & \SI{665.108}{mm}                 & Cameras' focal length                     \\
    FOV                       & $60.0^\circ$                     & Cameras' vertical FOV                     \\
    $\sigma^{\text{gyro}}$    & \SI{1.2e-4}{rad.s^{-1}\sqrt{Hz}} & Gyroscope noise density                   \\
    $\sigma_b^{\text{gyro}}$  & \SI{4.7e-6}{rad.s^{-2}\sqrt{Hz}} & Gyroscope random walk bias                \\
    $\sigma^{\text{accel}}$   & \SI{2.0e-3}{m.s^{-2}\sqrt{Hz}}   & Accelerometer noise density               \\
    $\sigma_b^{\text{accel}}$ & \SI{4.4e-5}{m.s^{-3}\sqrt{Hz}}   & Accelerometer random walk bias            \\
    $C_T$                     & \SI{2.27e-8}{N/rpm^2}            & RPM to thrust coefficient                 \\ 
    Image Size                & $1024~\text{px}\times768~\text{px}$                  & Image width and height                    \\
    \hline
\end{tabular}}
\vspace{-1cm}
  \vspace{1em}
\end{table}

\myparagraph{Experimental Setup}
Flights were performed in an $\met{11}\times\met{11}\times\met{5.5}$ motion capture room, with 24 OptiTrack Prime 17W cameras providing the 6D pose of the drone at \Hz{360}.
Each flight in the dataset is between 3-4 minutes long as the drone traces out a pre-defined periodic trajectory using a non-linear dynamic inversion controller \cite{talkaramanNDI}.
The drone is controlled and data is recorded by a custom software framework \cite{mccord2018ICRA} using the Lightweight Communications and Marshalling (LCM) protocol \cite{huang2010lcm}.

\myparagraph{Visual Data Generation}
Visual data was generated in post process using the FlightGoggles photo-realistic image generation system \cite{mccord2018ICRA}. FlightGoggles uses the ground truth 6D pose of the drone from motion capture to generate images from the viewpoint of each camera on the drone in a virtual environment.
The system allows for complete control over the visual appearance of the environment, the rate of camera images (up to the \Hz{360} motion capture rate), the number of cameras, and each camera's location and intrinsic and extrinsic properties.
The visual data generated by FlightGoggles has been previously validated for use in visual inertial state estimation in \cite{mccord2018ICRA}.

As part of the rendering process, a number of transforms are used to transform NED ground truth data from world frame into FlightGoggles' environment frame. To ensure that all recorded flights in each trajectory takeoff from a common altitude and overlap in the XY plane, $T_{mocap}^{norm}$ is introduced as a per-flight translational offset applied to the ground truth data to correct for offsets introduced during dataset collection. $T_{norm}^{FG_{env}}$ is a per-trajectory common transform that positions flights into the FlightGoggles environment in a collision-free manner. The full transform chain from ground-truth coordinates to render coordinates is shown in equation \ref{eq:transform}.

\begin{equation}
  T^{FG_{env}} = T_{norm}^{FG_{env}} * T_{mocap}^{norm} * T^{mocap}
  \label{eq:transform}
\end{equation}
Where $T^{FG_{env}}$ is the render pose in FlightGoggles' virtual environment. 

\myparagraph{Sensor Calibration and Temporal Synchronization}
The Kalibr package \cite{kalibr} was used to find the noise characteristics of the IMU and the IMU-to-camera transform.
A 3 second period at rest is included in every flight to allow for initialization of the time varying IMU bias.
%The IMU bias is estimated for each run during a 3 second period before take off using the averaged IMU data and the quadrotor orientation from motion capture.
% The optical odometry sensors directly count the number of rotations of each motor, and were validated using an external optical tachometer.
Force and torque coefficients of the drone were found experimentally through measurements in a wind tunnel to obtain the relationship between motor speeds and vehicle dynamics.
Clock synchronization between motion capture data and on-board sensors was performed using a combination of clock estimation over gigabit ethernet and chrony \cite{MiroslavLichvar} over the wireless network, with an upper bounded offset of $\pm5$ms.
\section{Dataset Format}
\label{sec:dataset}

Each flight within the dataset contains timestamped values for the following: ground truth 6D pose of the UAV at \Hz{360}, IMU measurements at \Hz{100}, RPM measurements for each motor at $\sim$\Hz{190}, and three camera streams (forward facing stereo pair and downward facing) at \Hz{120}.
The data is provided as grayscale images, LCM logs, and ROS bags for easy use in typical pipelines.
Scripts and binaries necessary to re-render images using FlightGoggles at other rates (up to \Hz{360}) or camera parameters and configurations are available at \url{http://blackbird-dataset.mit.edu/}.

\begin{figure}[htb]
  \centering
\pgfkeys{/pgf/inner sep=0.6mm}
\begin{forest}
  for tree={
    font=\ttfamily,
    grow'=0,
    child anchor=west,
    parent anchor=south,
    anchor=west,
    calign=first,
    edge path={
      \noexpand\path [draw, \forestoption{edge}]
      (!u.south west) +(7.5pt,0) |- node[fill,inner sep=1.25pt] {} (.child anchor)\forestoption{edge label};
    },
    before typesetting nodes={
      if n=1
        {insert before={[,phantom]}}
        {}
    },
    fit=band,
    before computing xy={l=15pt},
  }
[BlackbirdDataset
  [renderConfigs.yaml]
  [fileIndex.csv]
  [downloaderUtility.py]
  [renderUtilities
    [blackbirdDatasetUtils.py]
    [flightGogglesUtils.py]
  ]
  [BlackbirdDatasetData
    [{<trajectoryName>}
      [{<yawType> in \{yawConstant, yawForward\}}
        [{maxSpeed<V> in \{0p5, 1p0, 2p0, ..., 7p0\}}
          [{<trajectoryName>\_maxSpeed<V>.\{bag, log\}}]
          [groundTruthFlightNormalizationOffset.csv]
          [groundTruthPoses.csv]
          [images
            [{Camera\_<k>\_<renderName>.tar, k in \{L, R, D\}}
              [{<utime>\_Camera\_<k>.png}]
            ]][videos
            [<renderName>.mp4]]
        ]
    	]
    ]
  ]
]
\end{forest}
\caption{Dataset file hierarchy.}
\label{fig:dataset_format}
\end{figure}
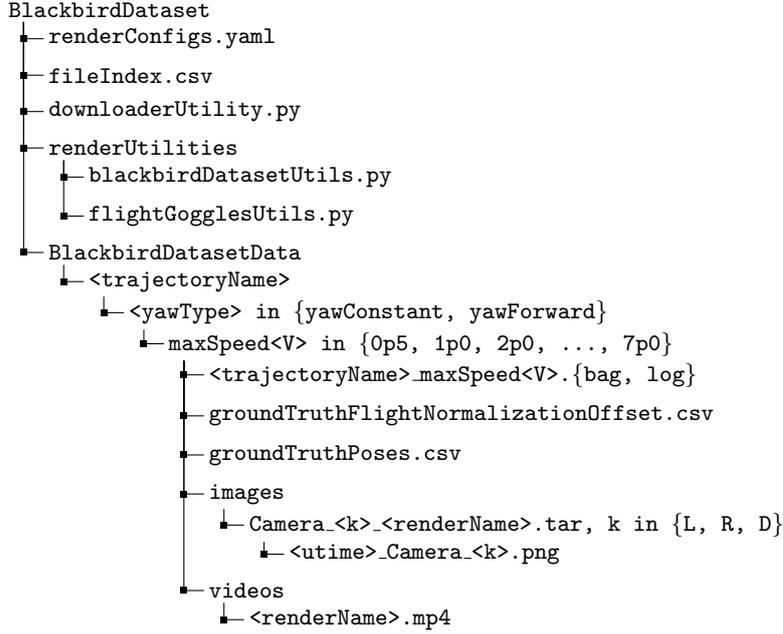

In addition to the raw data streams, the full calibration information of the UAV system is included in the dataset i.e, IMU noise characteristics, IMU-camera transform, camera intrinsic and extrinsic parameters (as currently rendered), and torque and thrust curves. The dataset's file structure is specified in figure \ref{fig:dataset_format}.

% 

% \wgcomment{Should I also add more CSV data here? Such as IMURaw.csv, camera timestamps, etc?}
\wgcomment{Revamp caption of file hierarchy and add the citation to text.}

\begin{figure}[hb]
  \vspace{-0.2in}
  \centering
  \hspace*{-0.6in}
  \begin{subfigure}{2.5in}
    \includegraphics[width=\textwidth]{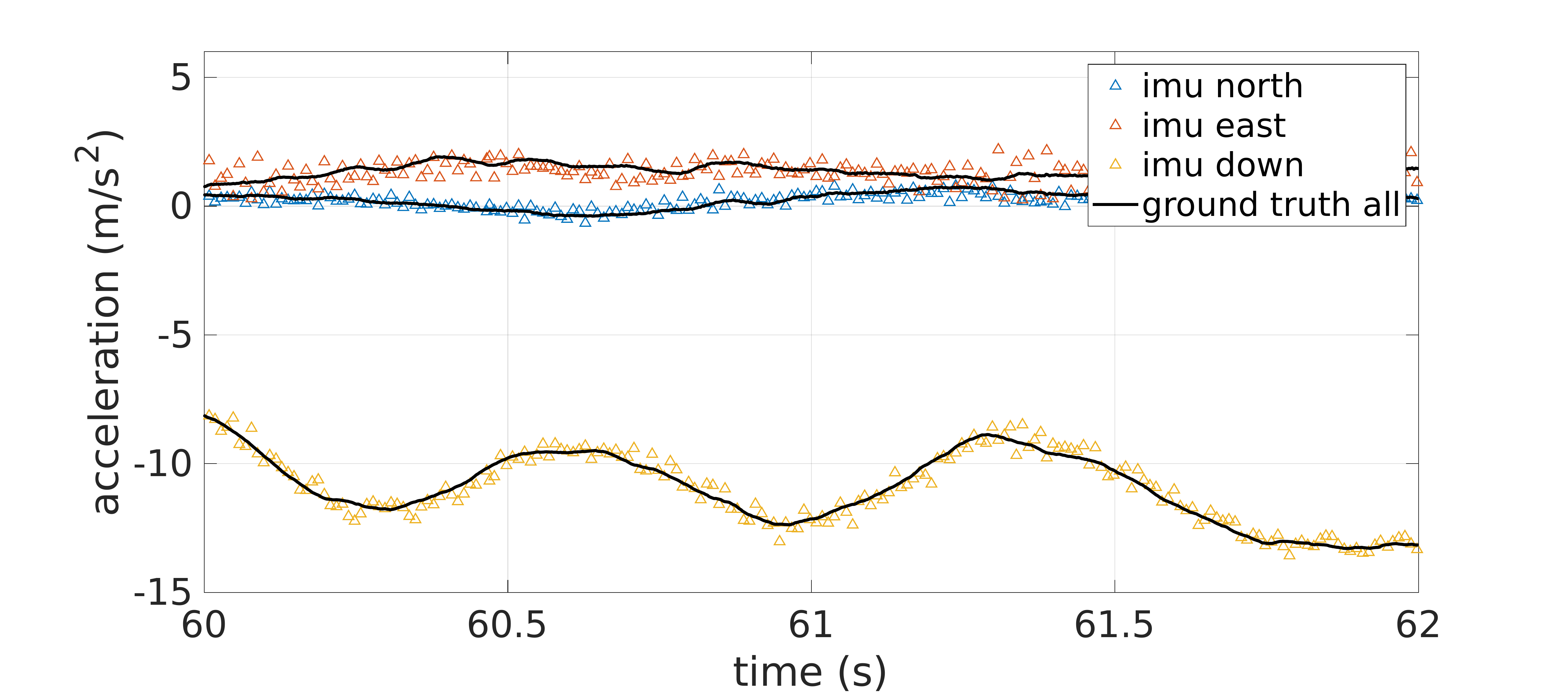}
    \caption{IMU accelerometer vs ground truth.}
    % \label{fig:pos_tracking}
  \end{subfigure}
  %\hspace{1em}
  \begin{subfigure}{2.5in}
    \includegraphics[width=\textwidth]{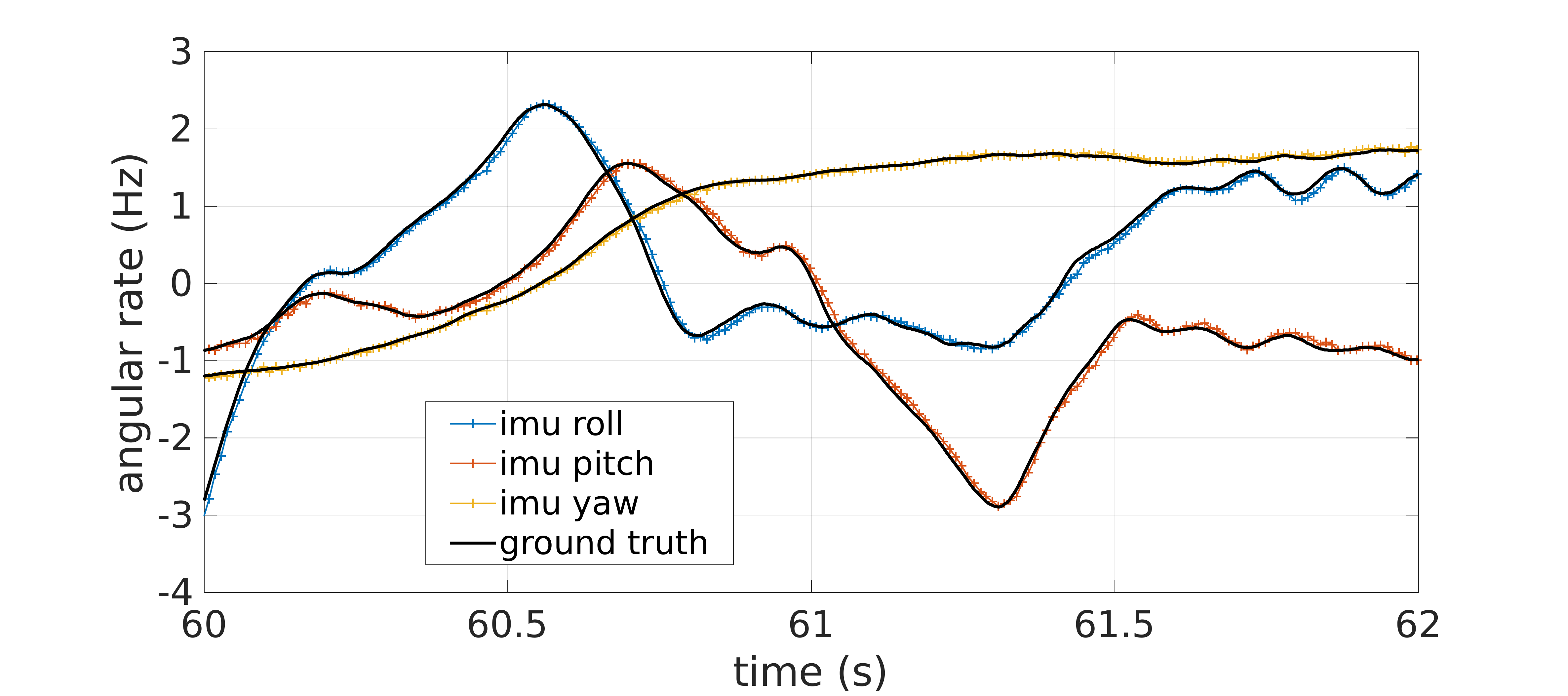}
    \caption{IMU gyroscope vs ground truth.}
    \label{fig:gyro_vs_ground_truth}
  \end{subfigure}
  \hspace*{-0.6in}
\caption{Derivative of position and rotational ground truth data compared with accelerometer and gyroscope data for a flight at \ms{4}.}
  \label{fig:mocap_accuracy}
  \vspace{-0.1in}
\end{figure}

%\begin{wrapfigure}{r}{0.5\textwidth}
% \begin{figure}[htb]
%   \vspace{-0.35in}
%   \centering
%   \hspace*{-.75\columnsep}
%   %\resizebox{0.5\textwidth}{!}{
%   \begin{subfigure}{2.5in}
%   \includegraphics[width=0.6\textwidth]{figures/gyro_vs_ground_truth.eps}
%   %}
%     \caption{Derivative of rotational ground truth data compared with gyroscope angular rate for a flight at \ms{4}.}
%   \label{fig:gyro_vs_ground_truth}
%   %\vspace{-0.25in}
% \end{figure}
%\end{wrapfigure}

\section{Data Validation}
\label{sec:dataValidation}
Validation of ground truth data and inertial measurements in both quality and temporal synchronization was performed by comparing raw inertial measurements with derivatives of the ground truth pose using a Savitzky-Golay filter \cite{savitzky1964smoothing}. \Cref{fig:mocap_accuracy} shows a comparison of the IMU angular rate and acceleration with respect to ground truth.
The accuracy of the drone's motor speed sensors were verified through the use of an external tachometer.

% In addition to ensuring accurate time synchronization between computers, all data is validated through careful analysis, which includes ascertaining that IMU measurements match the values computed from ground truth. We show this feature in \cref{fig:gyro_vs_ground_truth} by comparing the angular rate computed from ground truth information against the gyroscope measurements at a point where the speed reaches \ms{4}. The ground truth system accurately tracks the quadrotor even at highly dynamic maneuvers. In the plot, the angular rates from the IMU are corrected for gyroscope bias, which is estimated at the start of each trajectory before takeoff using ground truth information. We emphasize that this is an important feature which differentiates this dataset from previous \vmcomment{ones}{This sentence needs work!}.

\begin{figure}[h]
  \vspace{-0.2in}
  \centering
  \hspace*{-0.6in}
  \begin{subfigure}{2.5in}
    \includegraphics[width=\textwidth]{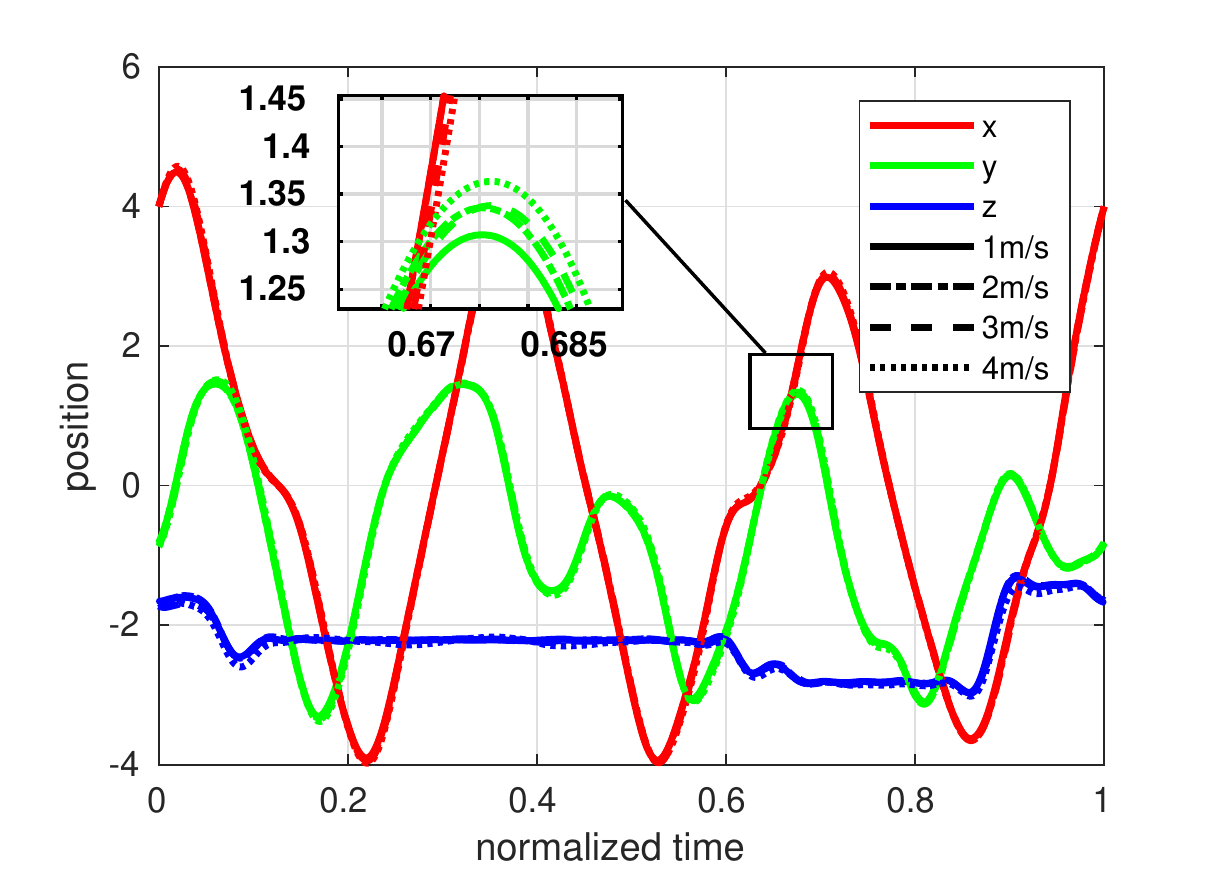}
    \caption{Position tracking}
    \label{fig:pos_tracking}
  \end{subfigure}
  %\hspace{1em}
  \begin{subfigure}{2.5in}
    \includegraphics[width=\textwidth]{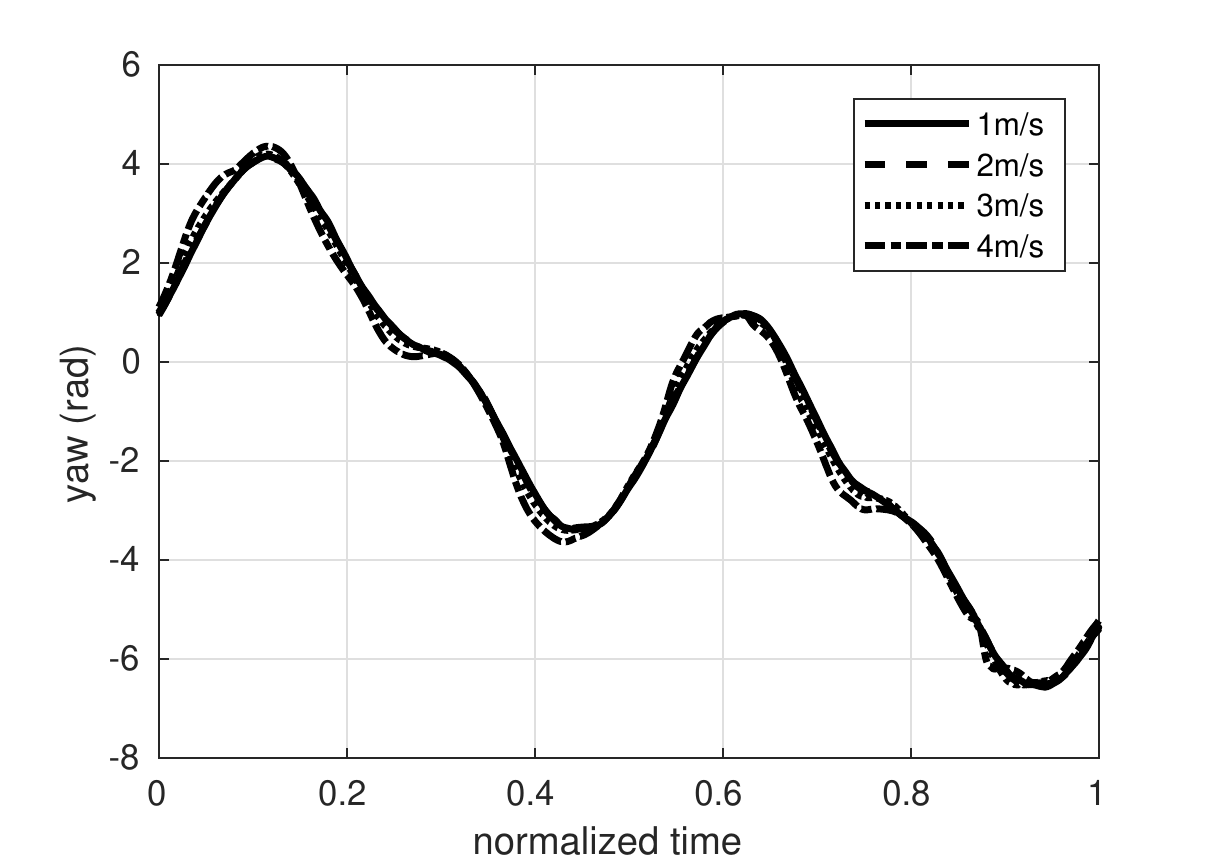}
    \caption{Yaw tracking}
    \label{fig:yaw_tracking}
  \end{subfigure}
  \hspace*{-0.6in}
\caption{Tracking precision while flying the same trajectory at speeds of \numrange{1}{4}\si{\meter\per\second}}
  \label{fig:tracking_accuracy}
  %\vspace{-0.1in}
\end{figure}

\myparagraph{Trajectory Tracking}
A feature of the provided dataset is the ability to repeatably run perception algorithms on a nominal trajectory pattern while flying at different speeds with new inertial, dynamical, and visual data.
A comparison of the ground truth pose of the same trajectory (Sphinx, see \cref{fig:trajectories}) flown four times at speeds between \ms{1} and \ms{4} is shown in Figure \ref{fig:tracking_accuracy}, with sufficient tracking accuracy for a user to isolate the speed of flight from other parameters that may affect VI-SLAM algorithms.

\myparagraph{Collision Checking}
To verify that the trajectories being rendered do not collide with any obstacles in the virtual environment, bullet3 \cite{coumans2015bullet} based simulation was used to verify that the trajectories were collision free prior to rendering the camera streams.

% The repeatability of trajectories at different speeds is demonstrated in \cref{fig:tracking_accuracy}, where the pose of the quadrotor is plotted for one full loop of the Sphinx trajectory at four speeds between \ms{1} and \ms{4}.

% Experiments completed or scheduled (one page)
\begin{table}[H]
\centering
\caption[Dataset Flights]{Dataset Flights}
\label{tab:experiments_completed}
\resizebox{0.9\textwidth}{!}{

\begin{tabular}{|c|cccccccc|cccccccc|}
\hline
 Trajectory & \multicolumn{8}{c|}{Constant Yaw} & \multicolumn{8}{c|}{Forward-Facing} \\ \hline
Top speed (\ms{}) & 0.5 & 1.0 & 2.0 & 3.0 & 4.0 & 5.0 & 6.0 & 7.0 & 0.5 & 1.0 & 2.0 & 3.0 & 4.0 & 5.0 & 6.0 & 7.0 \\ \hline
3D Figure 8 & \check & \check & \check & \check & \check & \check & - & - & - & - & - & - & - & - & - & -\tabularnewline
Ampersand & - & \href{http://blackbird-dataset.mit.edu/BlackbirdDatasetData/ampersand/yawConstant/maxSpeed1p0/videos/}{M} & \href{http://blackbird-dataset.mit.edu/BlackbirdDatasetData/ampersand/yawConstant/maxSpeed2p0/videos/}{M} & \href{http://blackbird-dataset.mit.edu/BlackbirdDatasetData/ampersand/yawConstant/maxSpeed3p0/videos/}{H} & - & - & - & - & - & \href{http://blackbird-dataset.mit.edu/BlackbirdDatasetData/ampersand/yawForward/maxSpeed1p0/videos/}{H} & \href{http://blackbird-dataset.mit.edu/BlackbirdDatasetData/ampersand/yawForward/maxSpeed2p0/videos/}{H} & - & - & - & - & -\tabularnewline
Bent Dice & - & \href{http://blackbird-dataset.mit.edu/BlackbirdDatasetData/bentDice/yawConstant/maxSpeed1p0/videos/}{E} & \href{http://blackbird-dataset.mit.edu/BlackbirdDatasetData/bentDice/yawConstant/maxSpeed2p0/videos/}{E} & \href{http://blackbird-dataset.mit.edu/BlackbirdDatasetData/bentDice/yawConstant/maxSpeed3p0/videos/}{M} & \href{http://blackbird-dataset.mit.edu/BlackbirdDatasetData/bentDice/yawConstant/maxSpeed4p0/videos/}{M} & - & - & - & \href{http://blackbird-dataset.mit.edu/BlackbirdDatasetData/bentDice/yawForward/maxSpeed0p5/videos/}{E} & \href{http://blackbird-dataset.mit.edu/BlackbirdDatasetData/bentDice/yawForward/maxSpeed1p0/videos/}{E} & \href{http://blackbird-dataset.mit.edu/BlackbirdDatasetData/bentDice/yawForward/maxSpeed2p0/videos/}{E} & \href{http://blackbird-dataset.mit.edu/BlackbirdDatasetData/bentDice/yawForward/maxSpeed3p0/videos/}{E} & - & - & - & -\tabularnewline
Clover & - & \href{http://blackbird-dataset.mit.edu/BlackbirdDatasetData/clover/yawConstant/maxSpeed1p0/videos/}{H} & \href{http://blackbird-dataset.mit.edu/BlackbirdDatasetData/clover/yawConstant/maxSpeed2p0/videos/}{H} & \href{http://blackbird-dataset.mit.edu/BlackbirdDatasetData/clover/yawConstant/maxSpeed3p0/videos/}{H} & \href{http://blackbird-dataset.mit.edu/BlackbirdDatasetData/clover/yawConstant/maxSpeed4p0/videos/}{H} & \href{http://blackbird-dataset.mit.edu/BlackbirdDatasetData/clover/yawConstant/maxSpeed5p0/videos/}{H} & \href{http://blackbird-dataset.mit.edu/BlackbirdDatasetData/clover/yawConstant/maxSpeed6p0/videos/}{H} & - & \href{http://blackbird-dataset.mit.edu/BlackbirdDatasetData/clover/yawForward/maxSpeed0p5/videos/}{H} & \href{http://blackbird-dataset.mit.edu/BlackbirdDatasetData/clover/yawForward/maxSpeed1p0/videos/}{H} & \href{http://blackbird-dataset.mit.edu/BlackbirdDatasetData/clover/yawForward/maxSpeed2p0/videos/}{H} & \href{http://blackbird-dataset.mit.edu/BlackbirdDatasetData/clover/yawForward/maxSpeed3p0/videos/}{H} & \href{http://blackbird-dataset.mit.edu/BlackbirdDatasetData/clover/yawForward/maxSpeed4p0/videos/}{H} & \href{http://blackbird-dataset.mit.edu/BlackbirdDatasetData/clover/yawForward/maxSpeed5p0/videos/}{H} & - & -\tabularnewline
Dice & - & - & \href{http://blackbird-dataset.mit.edu/BlackbirdDatasetData/dice/yawConstant/maxSpeed2p0/videos/}{E} & \href{http://blackbird-dataset.mit.edu/BlackbirdDatasetData/dice/yawConstant/maxSpeed3p0/videos/}{E} & \href{http://blackbird-dataset.mit.edu/BlackbirdDatasetData/dice/yawConstant/maxSpeed4p0/videos/}{M} & - & - & - & - & \href{http://blackbird-dataset.mit.edu/BlackbirdDatasetData/dice/yawForward/maxSpeed1p0/videos/}{E} & \href{http://blackbird-dataset.mit.edu/BlackbirdDatasetData/dice/yawForward/maxSpeed2p0/videos/}{E} & \href{http://blackbird-dataset.mit.edu/BlackbirdDatasetData/dice/yawForward/maxSpeed3p0/videos/}{E} & - & - & - & -\tabularnewline
Flat Figure 8 & \check & \check & \check & \check & - & \check & - & - & - & - & - & - & - & - & - & -\tabularnewline
Half-Moon & - & \href{http://blackbird-dataset.mit.edu/BlackbirdDatasetData/halfMoon/yawConstant/maxSpeed1p0/videos/}{E} & \href{http://blackbird-dataset.mit.edu/BlackbirdDatasetData/halfMoon/yawConstant/maxSpeed2p0/videos/}{E} & \href{http://blackbird-dataset.mit.edu/BlackbirdDatasetData/halfMoon/yawConstant/maxSpeed3p0/videos/}{E} & \href{http://blackbird-dataset.mit.edu/BlackbirdDatasetData/halfMoon/yawConstant/maxSpeed4p0/videos/}{M} & - & - & - & - & \href{http://blackbird-dataset.mit.edu/BlackbirdDatasetData/halfMoon/yawForward/maxSpeed1p0/videos/}{M} & \href{http://blackbird-dataset.mit.edu/BlackbirdDatasetData/halfMoon/yawForward/maxSpeed2p0/videos/}{M} & \href{http://blackbird-dataset.mit.edu/BlackbirdDatasetData/halfMoon/yawForward/maxSpeed3p0/videos/}{M} & \href{http://blackbird-dataset.mit.edu/BlackbirdDatasetData/halfMoon/yawForward/maxSpeed4p0/videos/}{M} & - & - & -\tabularnewline
Mouse & - & \href{http://blackbird-dataset.mit.edu/BlackbirdDatasetData/mouse/yawConstant/maxSpeed1p0/videos/}{M} & \href{http://blackbird-dataset.mit.edu/BlackbirdDatasetData/mouse/yawConstant/maxSpeed2p0/videos/}{M} & \href{http://blackbird-dataset.mit.edu/BlackbirdDatasetData/mouse/yawConstant/maxSpeed3p0/videos/}{M} & \href{http://blackbird-dataset.mit.edu/BlackbirdDatasetData/mouse/yawConstant/maxSpeed4p0/videos/}{M} & \href{http://blackbird-dataset.mit.edu/BlackbirdDatasetData/mouse/yawConstant/maxSpeed5p0/videos/}{M} & \href{http://blackbird-dataset.mit.edu/BlackbirdDatasetData/mouse/yawConstant/maxSpeed6p0/videos/}{M} & \href{http://blackbird-dataset.mit.edu/BlackbirdDatasetData/mouse/yawConstant/maxSpeed7p0/videos/}{M} & \href{http://blackbird-dataset.mit.edu/BlackbirdDatasetData/mouse/yawForward/maxSpeed0p5/videos/}{M} & \href{http://blackbird-dataset.mit.edu/BlackbirdDatasetData/mouse/yawForward/maxSpeed1p0/videos/}{M} & \href{http://blackbird-dataset.mit.edu/BlackbirdDatasetData/mouse/yawForward/maxSpeed2p0/videos/}{M} & \href{http://blackbird-dataset.mit.edu/BlackbirdDatasetData/mouse/yawForward/maxSpeed3p0/videos/}{M} & \href{http://blackbird-dataset.mit.edu/BlackbirdDatasetData/mouse/yawForward/maxSpeed4p0/videos/}{M} & \href{http://blackbird-dataset.mit.edu/BlackbirdDatasetData/mouse/yawForward/maxSpeed5p0/videos/}{M} & \href{http://blackbird-dataset.mit.edu/BlackbirdDatasetData/mouse/yawForward/maxSpeed6p0/videos/}{M} & \href{http://blackbird-dataset.mit.edu/BlackbirdDatasetData/mouse/yawForward/maxSpeed7p0/videos/}{M}\tabularnewline
Oval & - & - & \href{http://blackbird-dataset.mit.edu/BlackbirdDatasetData/oval/yawConstant/maxSpeed2p0/videos/}{M} & \href{http://blackbird-dataset.mit.edu/BlackbirdDatasetData/oval/yawConstant/maxSpeed3p0/videos/}{M} & \href{http://blackbird-dataset.mit.edu/BlackbirdDatasetData/oval/yawConstant/maxSpeed4p0/videos/}{H} & - & - & - & - & \href{http://blackbird-dataset.mit.edu/BlackbirdDatasetData/oval/yawForward/maxSpeed1p0/videos/}{M} & \href{http://blackbird-dataset.mit.edu/BlackbirdDatasetData/oval/yawForward/maxSpeed2p0/videos/}{M} & \href{http://blackbird-dataset.mit.edu/BlackbirdDatasetData/oval/yawForward/maxSpeed3p0/videos/}{H} & \href{http://blackbird-dataset.mit.edu/BlackbirdDatasetData/oval/yawForward/maxSpeed4p0/videos/}{H} & - & - & -\tabularnewline
Patrick & - & \href{http://blackbird-dataset.mit.edu/BlackbirdDatasetData/patrick/yawConstant/maxSpeed1p0/videos/}{E} & \href{http://blackbird-dataset.mit.edu/BlackbirdDatasetData/patrick/yawConstant/maxSpeed2p0/videos/}{E} & \href{http://blackbird-dataset.mit.edu/BlackbirdDatasetData/patrick/yawConstant/maxSpeed3p0/videos/}{E} & \href{http://blackbird-dataset.mit.edu/BlackbirdDatasetData/patrick/yawConstant/maxSpeed4p0/videos/}{E} & \href{http://blackbird-dataset.mit.edu/BlackbirdDatasetData/patrick/yawConstant/maxSpeed5p0/videos/}{M} & - & - & \href{http://blackbird-dataset.mit.edu/BlackbirdDatasetData/patrick/yawForward/maxSpeed0p5/videos/}{E} & \href{http://blackbird-dataset.mit.edu/BlackbirdDatasetData/patrick/yawForward/maxSpeed1p0/videos/}{E} & \href{http://blackbird-dataset.mit.edu/BlackbirdDatasetData/patrick/yawForward/maxSpeed2p0/videos/}{E} & \href{http://blackbird-dataset.mit.edu/BlackbirdDatasetData/patrick/yawForward/maxSpeed3p0/videos/}{E} & \href{http://blackbird-dataset.mit.edu/BlackbirdDatasetData/patrick/yawForward/maxSpeed4p0/videos/}{E} & - & - & -\tabularnewline
Picasso & \href{http://blackbird-dataset.mit.edu/BlackbirdDatasetData/picasso/yawConstant/maxSpeed0p5/videos/}{M} & \href{http://blackbird-dataset.mit.edu/BlackbirdDatasetData/picasso/yawConstant/maxSpeed1p0/videos/}{M} & \href{http://blackbird-dataset.mit.edu/BlackbirdDatasetData/picasso/yawConstant/maxSpeed2p0/videos/}{M} & \href{http://blackbird-dataset.mit.edu/BlackbirdDatasetData/picasso/yawConstant/maxSpeed3p0/videos/}{M} & \href{http://blackbird-dataset.mit.edu/BlackbirdDatasetData/picasso/yawConstant/maxSpeed4p0/videos/}{M} & \href{http://blackbird-dataset.mit.edu/BlackbirdDatasetData/picasso/yawConstant/maxSpeed5p0/videos/}{M} & \href{http://blackbird-dataset.mit.edu/BlackbirdDatasetData/picasso/yawConstant/maxSpeed6p0/videos/}{M} & - & \href{http://blackbird-dataset.mit.edu/BlackbirdDatasetData/picasso/yawForward/maxSpeed0p5/videos/}{M} & \href{http://blackbird-dataset.mit.edu/BlackbirdDatasetData/picasso/yawForward/maxSpeed1p0/videos/}{M} & - & \href{http://blackbird-dataset.mit.edu/BlackbirdDatasetData/picasso/yawForward/maxSpeed3p0/videos/}{H} & \href{http://blackbird-dataset.mit.edu/BlackbirdDatasetData/picasso/yawForward/maxSpeed4p0/videos/}{H} & \href{http://blackbird-dataset.mit.edu/BlackbirdDatasetData/picasso/yawForward/maxSpeed5p0/videos/}{H} & - & -\tabularnewline
Sid & - & \href{http://blackbird-dataset.mit.edu/BlackbirdDatasetData/sid/yawConstant/maxSpeed1p0/videos/}{E} & \href{http://blackbird-dataset.mit.edu/BlackbirdDatasetData/sid/yawConstant/maxSpeed2p0/videos/}{E} & \href{http://blackbird-dataset.mit.edu/BlackbirdDatasetData/sid/yawConstant/maxSpeed3p0/videos/}{E} & \href{http://blackbird-dataset.mit.edu/BlackbirdDatasetData/sid/yawConstant/maxSpeed4p0/videos/}{E} & \href{http://blackbird-dataset.mit.edu/BlackbirdDatasetData/sid/yawConstant/maxSpeed5p0/videos/}{E} & \href{http://blackbird-dataset.mit.edu/BlackbirdDatasetData/sid/yawConstant/maxSpeed6p0/videos/}{E} & \href{http://blackbird-dataset.mit.edu/BlackbirdDatasetData/sid/yawConstant/maxSpeed7p0/videos/}{E} & \href{http://blackbird-dataset.mit.edu/BlackbirdDatasetData/sid/yawForward/maxSpeed0p5/videos/}{M} & \href{http://blackbird-dataset.mit.edu/BlackbirdDatasetData/sid/yawForward/maxSpeed1p0/videos/}{M} & \href{http://blackbird-dataset.mit.edu/BlackbirdDatasetData/sid/yawForward/maxSpeed2p0/videos/}{M} & \href{http://blackbird-dataset.mit.edu/BlackbirdDatasetData/sid/yawForward/maxSpeed3p0/videos/}{M} & \href{http://blackbird-dataset.mit.edu/BlackbirdDatasetData/sid/yawForward/maxSpeed4p0/videos/}{M} & \href{http://blackbird-dataset.mit.edu/BlackbirdDatasetData/sid/yawForward/maxSpeed5p0/videos/}{M} & - & -\tabularnewline
Sphinx & - & \href{http://blackbird-dataset.mit.edu/BlackbirdDatasetData/sphinx/yawConstant/maxSpeed1p0/videos/}{H} & \href{http://blackbird-dataset.mit.edu/BlackbirdDatasetData/sphinx/yawConstant/maxSpeed2p0/videos/}{H} & \href{http://blackbird-dataset.mit.edu/BlackbirdDatasetData/sphinx/yawConstant/maxSpeed3p0/videos/}{H} & \href{http://blackbird-dataset.mit.edu/BlackbirdDatasetData/sphinx/yawConstant/maxSpeed4p0/videos/}{H} & - & - & - & - & \href{http://blackbird-dataset.mit.edu/BlackbirdDatasetData/sphinx/yawForward/maxSpeed1p0/videos/}{M} & \href{http://blackbird-dataset.mit.edu/BlackbirdDatasetData/sphinx/yawForward/maxSpeed2p0/videos/}{M} & \href{http://blackbird-dataset.mit.edu/BlackbirdDatasetData/sphinx/yawForward/maxSpeed3p0/videos/}{M} & \href{http://blackbird-dataset.mit.edu/BlackbirdDatasetData/sphinx/yawForward/maxSpeed4p0/videos/}{M} & - & - & -\tabularnewline
Star & - & \href{http://blackbird-dataset.mit.edu/BlackbirdDatasetData/star/yawConstant/maxSpeed1p0/videos/}{M} & \href{http://blackbird-dataset.mit.edu/BlackbirdDatasetData/star/yawConstant/maxSpeed2p0/videos/}{M} & \href{http://blackbird-dataset.mit.edu/BlackbirdDatasetData/star/yawConstant/maxSpeed3p0/videos/}{M} & \href{http://blackbird-dataset.mit.edu/BlackbirdDatasetData/star/yawConstant/maxSpeed4p0/videos/}{H} & \href{http://blackbird-dataset.mit.edu/BlackbirdDatasetData/star/yawConstant/maxSpeed5p0/videos/}{H} & - & - & \href{http://blackbird-dataset.mit.edu/BlackbirdDatasetData/star/yawForward/maxSpeed0p5/videos/}{M} & \href{http://blackbird-dataset.mit.edu/BlackbirdDatasetData/star/yawForward/maxSpeed1p0/videos/}{M} & \href{http://blackbird-dataset.mit.edu/BlackbirdDatasetData/star/yawForward/maxSpeed2p0/videos/}{M} & \href{http://blackbird-dataset.mit.edu/BlackbirdDatasetData/star/yawForward/maxSpeed3p0/videos/}{M} & \href{http://blackbird-dataset.mit.edu/BlackbirdDatasetData/star/yawForward/maxSpeed4p0/videos/}{M} & \href{http://blackbird-dataset.mit.edu/BlackbirdDatasetData/star/yawForward/maxSpeed5p0/videos/}{H} & - & -\tabularnewline
Thrice & - & \href{http://blackbird-dataset.mit.edu/BlackbirdDatasetData/thrice/yawConstant/maxSpeed1p0/videos/}{E} & \href{http://blackbird-dataset.mit.edu/BlackbirdDatasetData/thrice/yawConstant/maxSpeed2p0/videos/}{E} & \href{http://blackbird-dataset.mit.edu/BlackbirdDatasetData/thrice/yawConstant/maxSpeed3p0/videos/}{E} & \href{http://blackbird-dataset.mit.edu/BlackbirdDatasetData/thrice/yawConstant/maxSpeed4p0/videos/}{E} & \href{http://blackbird-dataset.mit.edu/BlackbirdDatasetData/thrice/yawConstant/maxSpeed5p0/videos/}{E} & \href{http://blackbird-dataset.mit.edu/BlackbirdDatasetData/thrice/yawConstant/maxSpeed6p0/videos/}{M} & \href{http://blackbird-dataset.mit.edu/BlackbirdDatasetData/thrice/yawConstant/maxSpeed7p0/videos/}{M} & \href{http://blackbird-dataset.mit.edu/BlackbirdDatasetData/thrice/yawForward/maxSpeed0p5/videos/}{E} & \href{http://blackbird-dataset.mit.edu/BlackbirdDatasetData/thrice/yawForward/maxSpeed1p0/videos/}{E} & \href{http://blackbird-dataset.mit.edu/BlackbirdDatasetData/thrice/yawForward/maxSpeed2p0/videos/}{E} & \href{http://blackbird-dataset.mit.edu/BlackbirdDatasetData/thrice/yawForward/maxSpeed3p0/videos/}{E} & \href{http://blackbird-dataset.mit.edu/BlackbirdDatasetData/thrice/yawForward/maxSpeed4p0/videos/}{E} & \href{http://blackbird-dataset.mit.edu/BlackbirdDatasetData/thrice/yawForward/maxSpeed5p0/videos/}{E} & \href{http://blackbird-dataset.mit.edu/BlackbirdDatasetData/thrice/yawForward/maxSpeed6p0/videos/}{M} & -\tabularnewline
Tilted Thrice & - & \href{http://blackbird-dataset.mit.edu/BlackbirdDatasetData/tiltedThrice/yawConstant/maxSpeed1p0/videos/}{E} & \href{http://blackbird-dataset.mit.edu/BlackbirdDatasetData/tiltedThrice/yawConstant/maxSpeed2p0/videos/}{E} & \href{http://blackbird-dataset.mit.edu/BlackbirdDatasetData/tiltedThrice/yawConstant/maxSpeed3p0/videos/}{E} & \href{http://blackbird-dataset.mit.edu/BlackbirdDatasetData/tiltedThrice/yawConstant/maxSpeed4p0/videos/}{E} & \href{http://blackbird-dataset.mit.edu/BlackbirdDatasetData/tiltedThrice/yawConstant/maxSpeed5p0/videos/}{E} & \href{http://blackbird-dataset.mit.edu/BlackbirdDatasetData/tiltedThrice/yawConstant/maxSpeed6p0/videos/}{M} & \href{http://blackbird-dataset.mit.edu/BlackbirdDatasetData/tiltedThrice/yawConstant/maxSpeed7p0/videos/}{M} & \href{http://blackbird-dataset.mit.edu/BlackbirdDatasetData/tiltedThrice/yawForward/maxSpeed0p5/videos/}{E} & \href{http://blackbird-dataset.mit.edu/BlackbirdDatasetData/tiltedThrice/yawForward/maxSpeed1p0/videos/}{E} & \href{http://blackbird-dataset.mit.edu/BlackbirdDatasetData/tiltedThrice/yawForward/maxSpeed2p0/videos/}{E} & \href{http://blackbird-dataset.mit.edu/BlackbirdDatasetData/tiltedThrice/yawForward/maxSpeed3p0/videos/}{E} & \href{http://blackbird-dataset.mit.edu/BlackbirdDatasetData/tiltedThrice/yawForward/maxSpeed4p0/videos/}{E} & \href{http://blackbird-dataset.mit.edu/BlackbirdDatasetData/tiltedThrice/yawForward/maxSpeed5p0/videos/}{E} & \href{http://blackbird-dataset.mit.edu/BlackbirdDatasetData/tiltedThrice/yawForward/maxSpeed6p0/videos/}{E} & -\tabularnewline
Winter & - & \href{http://blackbird-dataset.mit.edu/BlackbirdDatasetData/winter/yawConstant/maxSpeed1p0/videos/}{M} & \href{http://blackbird-dataset.mit.edu/BlackbirdDatasetData/winter/yawConstant/maxSpeed2p0/videos/}{M} & \href{http://blackbird-dataset.mit.edu/BlackbirdDatasetData/winter/yawConstant/maxSpeed3p0/videos/}{M} & \href{http://blackbird-dataset.mit.edu/BlackbirdDatasetData/winter/yawConstant/maxSpeed4p0/videos/}{M} & \href{http://blackbird-dataset.mit.edu/BlackbirdDatasetData/winter/yawConstant/maxSpeed5p0/videos/}{M} & - & - & \href{http://blackbird-dataset.mit.edu/BlackbirdDatasetData/winter/yawForward/maxSpeed0p5/videos/}{M} & - & \href{http://blackbird-dataset.mit.edu/BlackbirdDatasetData/winter/yawForward/maxSpeed2p0/videos/}{H} & \href{http://blackbird-dataset.mit.edu/BlackbirdDatasetData/winter/yawForward/maxSpeed3p0/videos/}{H} & \href{http://blackbird-dataset.mit.edu/BlackbirdDatasetData/winter/yawForward/maxSpeed4p0/videos/}{H} & - & - & -\tabularnewline
\hline
\end{tabular}
}\\% resizebox
Click flight for grayscale video preview of flight in all rendered environments. 
\end{table}

\section{Datasets}
\label{sec:Datasets}

\begin{figure}
  \centering
  \begin{subfigure}[t]{3.9cm}
    \centering
    \includegraphics[width=3.9cm]{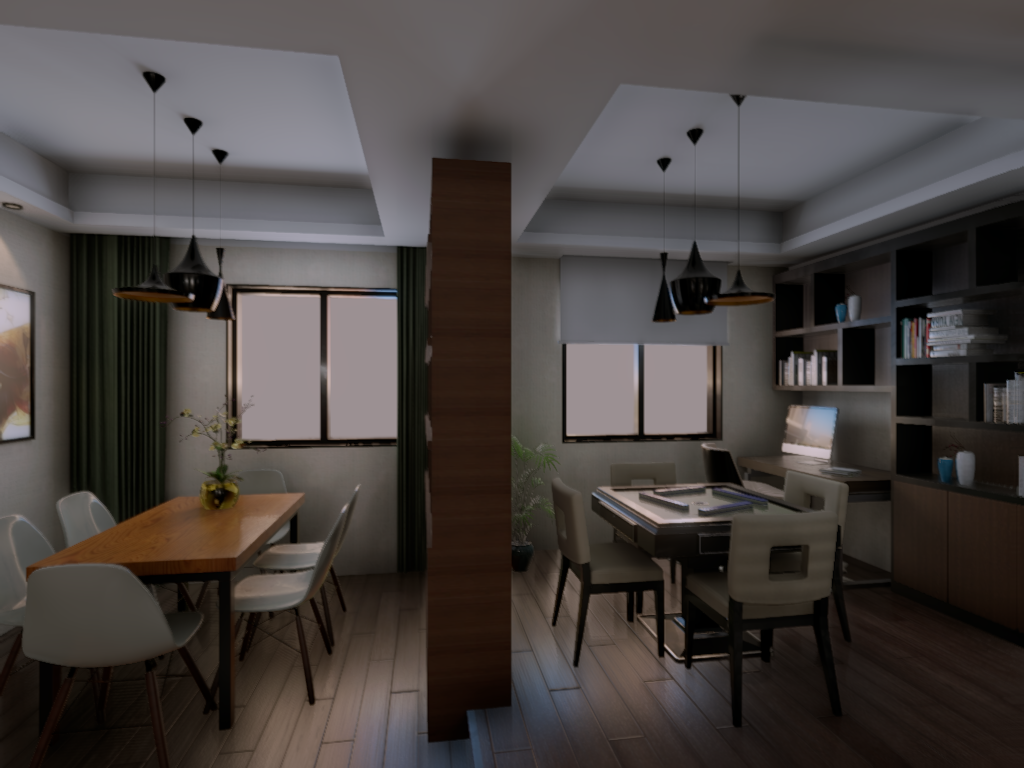}
    \vspace{-0.4cm}
    \subcaption{}
    \label{subfig:BA}
  \end{subfigure}
  \hfill
  \begin{subfigure}[t]{3.9cm}
    \centering
    \includegraphics[width=3.9cm]{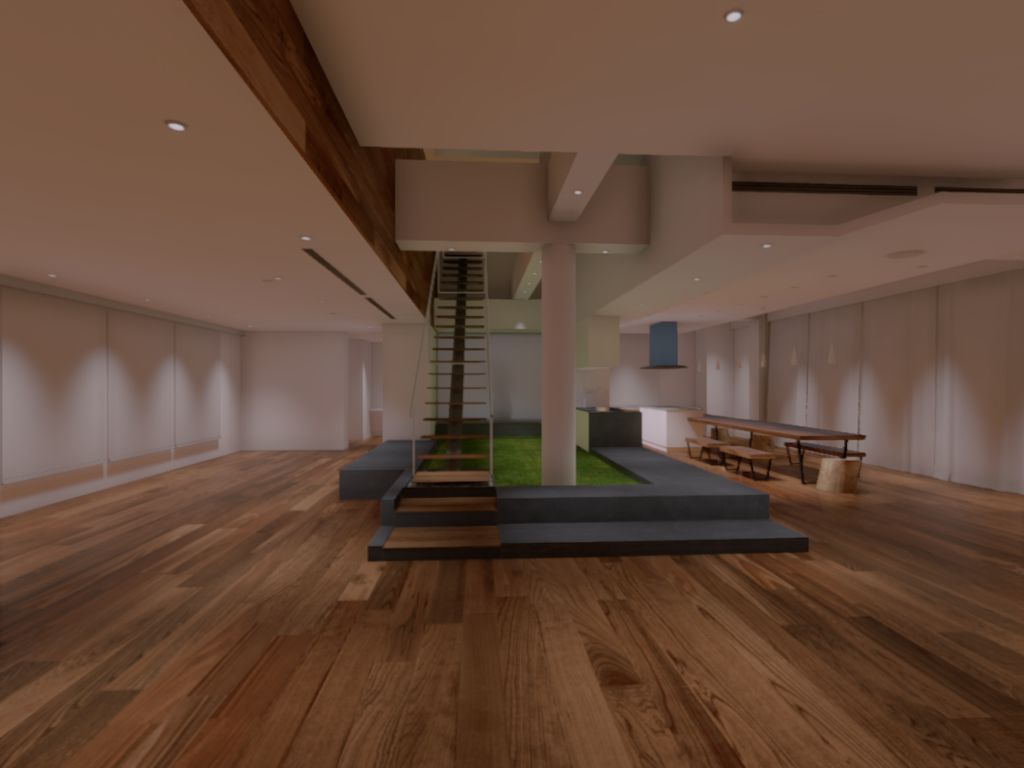}
    \vspace{-0.4cm}
    \subcaption{}
    \label{subfig:HL}
  \end{subfigure}
  \hfill
    \begin{subfigure}[t]{3.9cm}
    \centering
    \includegraphics[width=3.9cm]{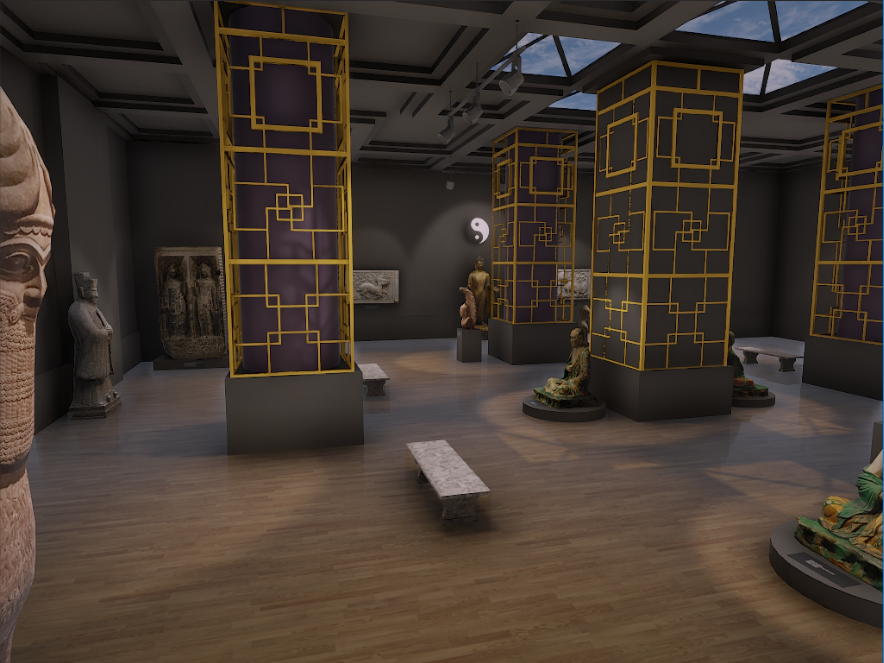}
    \vspace{-0.4cm}
    \subcaption{}
    \vspace{0.2cm}
    \label{subfig:MP}
  \end{subfigure}
  %\hfill
  \begin{subfigure}[t]{3.9cm}
    \centering
    \includegraphics[width=3.9cm]{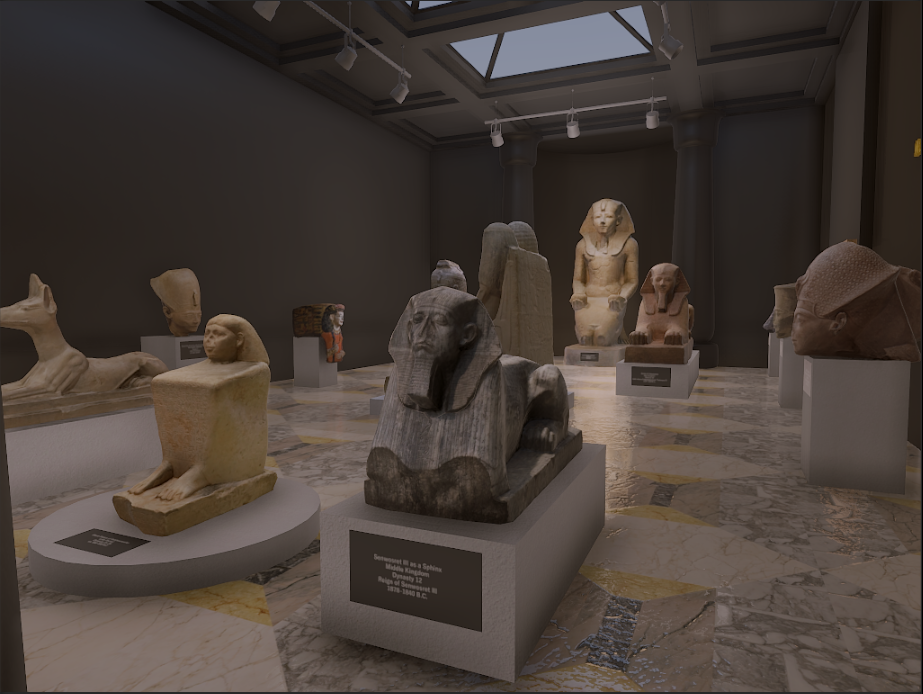}
    \vspace{-0.4cm}
    \subcaption{}
    \label{subfig:MS}
  \end{subfigure}
  \hspace{0.1cm}
    \begin{subfigure}[t]{3.9cm}
    \centering
    \includegraphics[width=3.9cm]{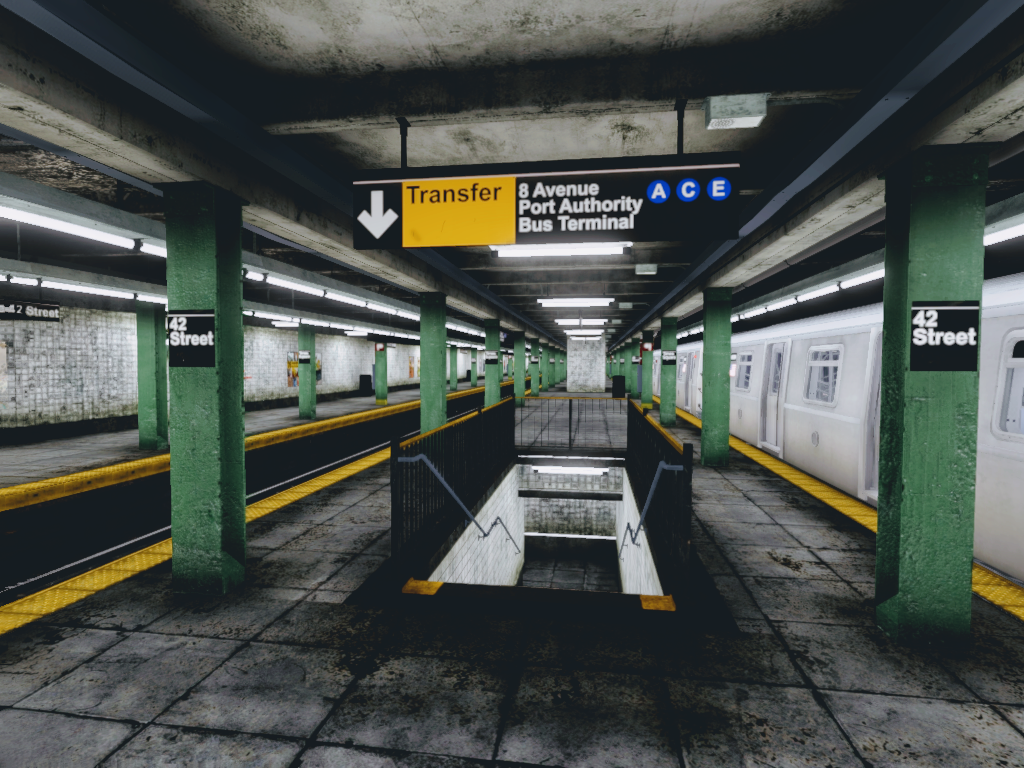}
    \vspace{-0.4cm}
    \subcaption{}
    \label{subfig:SW}
  \end{subfigure}
  %\hfill
  \vspace{-0.12in}
  \caption{Five rendering environments for visual data: (\subref{subfig:BA}) Butterfly Apartment, (\subref{subfig:HL}) Hazelwood Loft, (\subref{subfig:MP}) Museum Pillars, (\subref{subfig:MS}) Museum Sphinx, and (\subref{subfig:SW}) NYC Subway}
  \label{fig:environments}
  \vspace{-0.2cm}
\end{figure}

\begin{figure}
  \centering
  %\vspace{-0.1in}
  \includegraphics[width=\textwidth,trim={2cm, 1.2cm, 2cm, 0},clip]{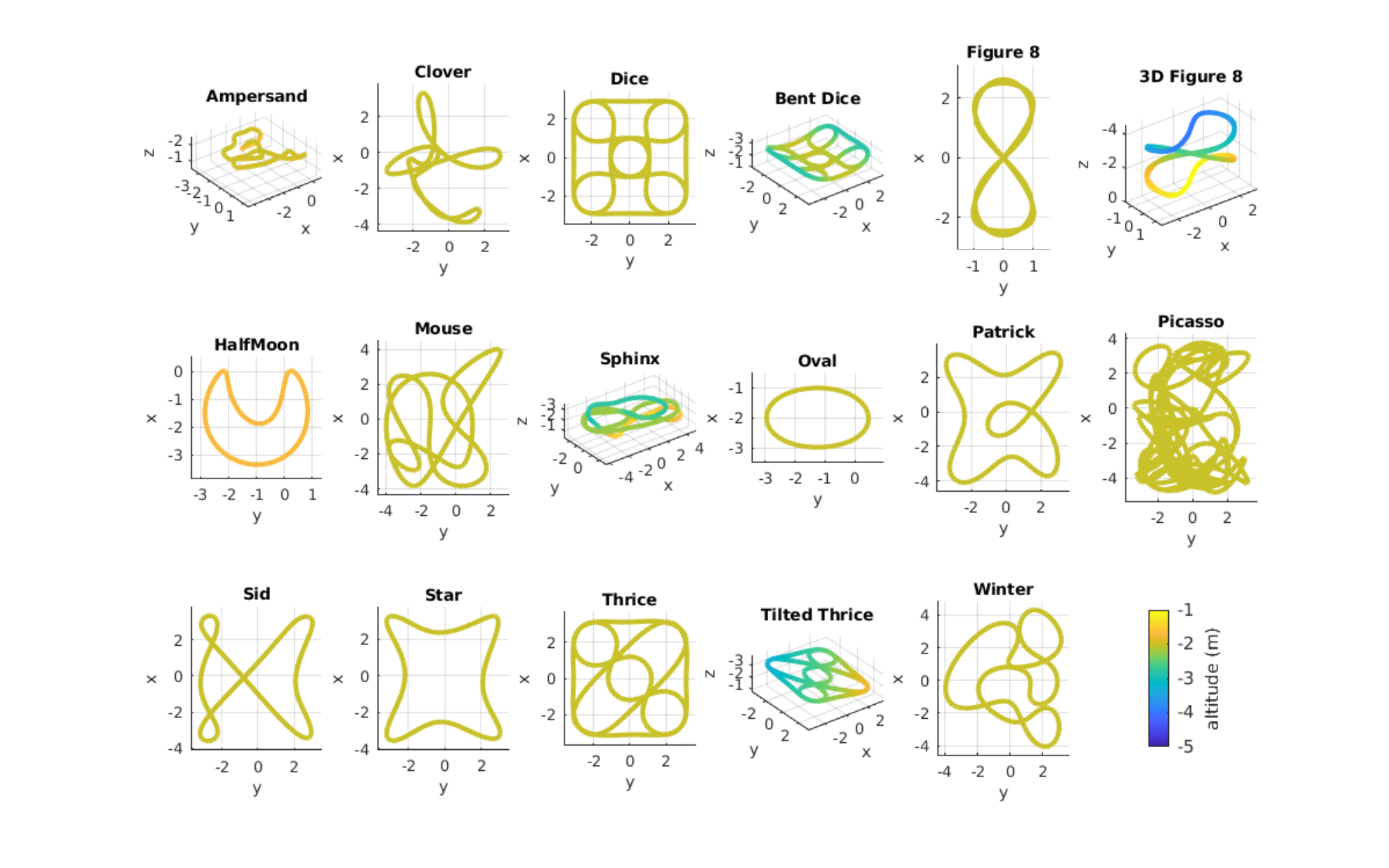}
  \caption{Top down diagrams of trajectory paths included in this dataset.}
  \label{fig:trajectories}
  \vspace{-0.7cm}
\end{figure}

The trajectories were designed to allow for independent variation in the following flight characteristics: speed, yaw, trajectory complexity, and period. The included trajectories are classified by difficulty ([E]asy, [M]edium, [H]ard) according to mean features tracked using the visual inertial state estimation pipeline outlined in \cite{mccord2018ICRA} and are shown in \cref{tab:experiments_completed}. These trajectories range in complexity from an oval with constant yaw and altitude to trajectories with varying speed, altitude, and yaw as they weave through visual obstacles (e.g. the Sphinx) as shown in \cref{fig:trajectories}. For flights that are rendered in multiple environments (see \cref{fig:environments}), some environments are harder for state estimation than others.

To generate smoothly trackable trajectories, minimum snap optimization was performed over a set of requested waypoint positions using the non-linear optimization technique from \cite{burri2015real-time}, with boundary conditions to make the trajectory periodic.
\Cref{tab:experiments_completed} shows all the sequences included. There are 163 unique flights of approximately three minutes each for a total of over \hr{10} and \SI{60}{\kilo\meter} of ground truth, inertial, and dynamical sensor data, as well as rendered imagery in multiple environments.
Trajectories were designed for specific flight environments (e.g. Sphinx for Museum Sphinx) and are therefore particularly well suited to those environments, however, where it does not result in virtual collisions with objects, the same trajectory can be re-rendered in multiple environments. Flights marked in \cref{tab:experiments_completed} by \check do not have visual data associated with them and are intended for calibration use.

% The Oval, Half-Moon, and Ampersand trajectories are designed to fit inside an apartment environment and are smaller than the rest. Therefore, they can easily run in a wide range of environments, including the four presented. These trajectories were flown, for each speed, with both forward-facing and zero yaw configurations. Among those, the forward-facing \ms{3} Ampersand is one of the most challenging ones for VI-SLAM in our experience. It has many rapid changes in velocity as well as near in-place. All 33 flights can be reproduced with different camera configurations.

\section{Known Issues}
\label{sec:known_issues}

% \begin{itemize}
In this dataset, we synchronized motion capture ground truth data with onboard IMU measurements using camera exposure timestamps provided by OptiTrack, IMU measurement timestamps provided on arrival by our UAV's TX2, and clock sync and drift correction provided by Chrony \cite{MiroslavLichvar}. However, due to the complexity and stochastic nature of the systems involved, we are only able to guarantee IMU and ground truth temporal alignment to within $\pm$5ms across all flights in this dataset. This upperbound was verified in post process by cross correlation of IMU and ground truth measurements for each flight. 

Over the course of the various data recording sessions required to create this dataset, many recalibrations of the motion capture groundtruth setup were required due to thermal expansion and contraction of the motion capture support beams. Each recalibration's overall tracking error and groundplane alignment was verified using post process analysis, but are not guaranteed to be exact. Most flights were flown with $\leq$1.5mm of mean tracking error and $\leq1^\circ$ of ground plane alignment error.  
    
% \input{paper_sections/calibration_and_synchronization.tex}

% \input{paper_sections/known_issues.tex}

% \input{paper_sections/data_access_methods.tex}

% Main Experimental Insights (one page)
\section{Conclusion}
\label{sec:main_experimental_insights}

The Blackbird dataset allows for the evaluation of the robustness and performance of VI-SLAM algorithms under varying conditions and degrees of agility. It has been used in \cite{AntoniniSM2018} to evaluate dynamics factors in factor graph VIO. As an additional example, we show the results of running a tracker-based visual inertial odometry method \cite{mccord2018ICRA} on the Ampersand trajectory for increasing speed of execution.
The performance is summarized in \Cref{tab:resultsVIO}, where drift rate (the percentage error per meter traveled) serves as a metric of the overall state estimation accuracy. The mean features tracked, the average number of features tracked between every pair of consecutive frames, serves as a metric of success for the visual front-end.
The structured nature of the dataset allows for clear evaluations of the strengths and failures of particular algorithms (e.g. tracking vs. matching visual front-ends) depending on the type of trajectory and visual environment being flown in.

% This analysis is shown in \Cref{tab:resultsVIO}. The drift rate is defined as the percentage of error per meter travelled. The mean features tracked is the mean of the number of features tracked between consecutive frames. This is indicative of the challenges for tracking based methods even at high framerates with increased agility.

\begin {table} [h]
\caption {Performance of Visual Inertial Odometry \cite{mccord2018ICRA} for the same nominal trajectory with varying speed}
	\begin {tabularx} {\textwidth} {XXX}
	\hline
	Speed & Drift Rate & Mean Features Tracked \tabularnewline
	\hline
	\ms{1.12} & 0.55 & 118.98 \tabularnewline
	\ms{1.96} & 0.95 & 115.74 \tabularnewline
	\ms{3.08} & 1.94 & 106.83 \tabularnewline
	\hline
	\end{tabularx}
\label{tab:resultsVIO}
\end{table}

% \myparagraph{Conclusion} 
% The dataset described here provides a large quantity of flight data for a UAV flying at high speeds performing agile maneuvers in an indoor environment with precise ground truth.
%

Due to the large number of aggressive and varied flight patterns available, the Blackbird dataset is well suited to fill a void in the current landscape of available robotics datasets. The inertial, dynamical, and visual sensors provide a complete sensor package for many UAV VI-SLAM algorithms.
The photorealistic visual simulation system, FlightGoggles, allows a user to expand beyond the three camera streams provided and re-render images to match the characteristics of their current system, or to evaluate design choices on camera placement, type, and framerate. %Additionally, FlightGoggles can provide ground truth depth images, which are useful for evaluating the accuracy of feature triangulaion and stereo reconstruction algorithms.
Variations in visual parameters may all be performed while maintaining the true dynamics and inertial measurements of the UAV, and without sacrificing the high resolution ground truth accuracy provided by an expensive motion capture system.

The ability to repeat nominal trajectories with varying conditions allows for methodical evaluation of perception algorithm performance during high speed flight and enables progressively building capabilities towards more challenging scenarios.

\bibliographystyle{splncs}
\bibliography{references}

\end{document}